\documentclass{article}
\usepackage{ijcai22}

\usepackage{times}
\usepackage{soul}
\usepackage{url}
\usepackage[hidelinks]{hyperref}
\usepackage[utf8]{inputenc}
\usepackage[small]{caption}
\usepackage{graphicx}
\usepackage{amsmath}
\usepackage{amsthm}
\usepackage{booktabs}
\usepackage{algorithm}
\usepackage{algorithmic}

\usepackage{tikz}
\usepackage{comment}
\usepackage{amsmath,amssymb} 
\usepackage{color}

\usepackage{graphicx}
\usepackage{amsmath}
\usepackage{amssymb}
\usepackage{booktabs}
\usepackage{amsmath}
\usepackage{tabularx}

\title{ReplaceBlock: An improved regularization method based on background information}

\author{
Zhemin Zhang$^1$
\and
Xun Gong$^1$\and
Jinyi Wu$^1$
\affiliations
$^1$Southwest Jiaotong University, China
\emails
zheminzhang@my.swjtu.edu.cn
}

\begin{document}

\maketitle

\begin{abstract}
Attention mechanism, being frequently used to train networks for better feature representations, can effectively disentangle the target object from irrelevant objects in the background. Given an arbitrary image, we find that the background's irrelevant objects are most likely to occlude/block the target object. We propose, based on this finding, a ReplaceBlock to simulate the situations when the target object is partially occluded by the objects that are deemed as background. Specifically, ReplaceBlock erases the target object in the image, and then generates a feature map with only irrelevant objects and background by the model. Finally, some regions in the background feature map are used to replace some regions of the target object in the original image feature map. In this way, ReplaceBlock can effectively simulate the feature map of the occluded image. The experimental results show that ReplaceBlock works better than DropBlock in regularizing convolutional networks.
\end{abstract}

\section{Introduction}

Convolutional neural networks (CNNs) have been widely and successfully used in various computer vision tasks \cite{krizhevsky2012imagenet,he2017mask}. In recent years, various deep CNN architectures have significantly improved the accuracy of image classification, such as ResNet \cite{he2016deep} and InceptionNet \cite{szegedy2016rethinking}. Deep CNNs usually have a large number of parameters, which can very easily lead to over-fitting of the network. Therefore, it is worthy of developing regularization methods to alleviate the problem of over-ﬁtting for CNNs. However, a recent study shows that traditional Dropout is less effective for CNNs because of the under-drop problem \cite{ghiasi2018dropblock,zeng2020corrdrop}. To address this issue, DropBlock \cite{ghiasi2018dropblock} developed a structured form of Dropout for CNN models, where contiguous regions of feature maps are dropped instead of individual elements in the feature maps.

\begin{figure}[t]
  \centering
   \includegraphics[width=1.0\linewidth]{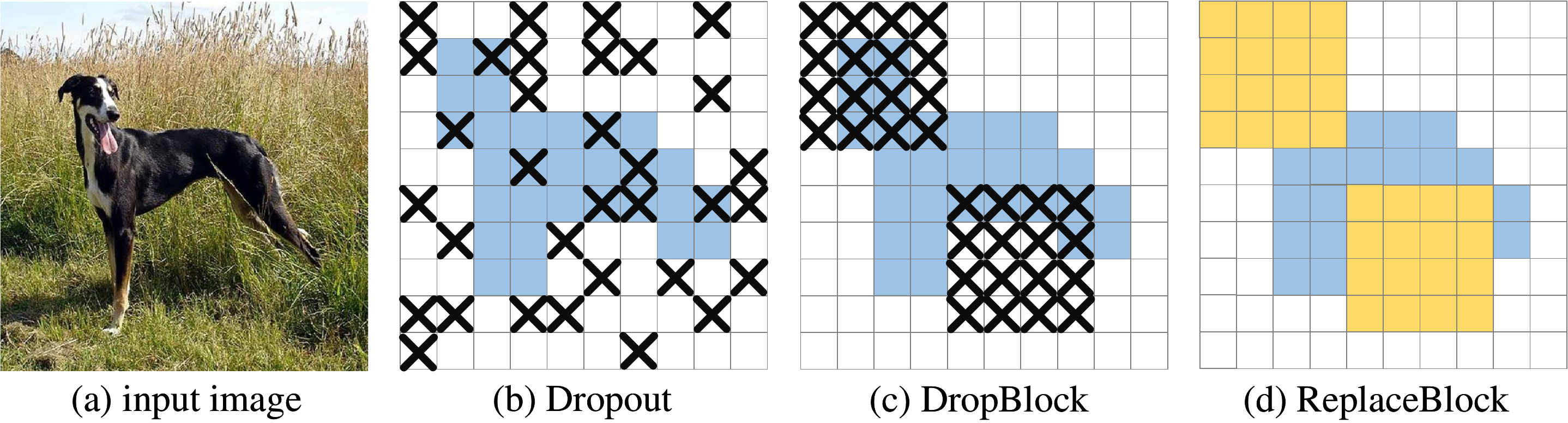}

   \caption{Masks of Dropout, DropBlock and our ReplaceBlock. The blue regions in (b), (c) and (d) represents the region where the input image is highly discriminative. The yellow regions in (d) indicate the depth feature map blocks of the background information, as shown in Figure \ref{MethodBlockDiagram}. This figure follows DropBlock.}
   \label{dropBlockVSreplaceBlock}
\end{figure}

Apart from the regularization approaches, we investigate the effect of attention in CNNs. The signiﬁcance of attention has been studied extensively in the previous literature \cite{jaderberg2015spatial,woo2018cbam,yang2021simam}. The attention mechanism can distinguish between the target object and background of an image, which is widely used in weakly supervised object localization (WSOL) \cite{zhou2016learning,choe2019attention,yin2021dual}. WSOL aims to identify the object's location in a scene only using image-level labels, not location annotations. This is consistent with our goal of using attention, only using image-level labels, to obtain an image with only the background (hereinafter called \textbf{background-only image}) through the attention mechanism.

To date, the general practice of regularization methods related to Dropout \cite{srivastava2014dropout} has been to drop some of the features on the feature map directly, i.e., to erase the contiguous regions or individual pixels on the feature maps by setting them zero during the training phase, to encourage the models look elsewhere for evidence to fit the data. In WSOL, Dropout is usually combined with attention mechanism \cite{choe2019attention,yin2021dual}, erasing the most discriminative region on the feature map by zeroing that region during the training phase. This prevents the model from relying solely on the most discriminative part for classiﬁcation and encourages it to learn the less discriminative part as well. It was demonstrated in CamDrop \cite{10.1145/3357384.3357999} that erasing the most discriminative part is effective to capture the full extent of object and obtain a more accurate object localization. However, none of these methods can simulate the occluded image in a natural way. Because in the real world, it is usually a specific object that occludes the target object, not a 0-1 mask. Although the attention mechanism can distinguish the target object and background of an image, unfortunately, the current attention technique rarely utilizes the information of the background region of an image.

From previous methods, we conclude that erasing the contiguous regions of the feature map can effectively alleviate the over-fitting of the model. But unlike the previous approach of direct dropping, in this paper, we propose a novel and efﬁcient regularization method, ReplaceBlock, as shown in Figure \ref{dropBlockVSreplaceBlock}. Utilized the ability of the attention mechanism to distinguish the target object and background of input images, we can obtain feature maps with only background information. ReplaceBlock replaces a small region of the target object in the original image feature map with background-only feature map blocks to simulate the feature map where the target object in the image is partially occluded. Specifically, a spatial self-attention map is obtained from the final feature map of the network. Based on the spatial self-attention map, we produce the first component of the ReplaceBlock, i.e., the target class object drop mask (\textbf{TC-DM}). We obtain this TC-DM by thresholding the spatial self-attention map. During the training, we first use the TC-DM to hide the objects to be recognized in the original image, and then use the model to obtain a deep feature map with only irrelevant background information. Also based on the spatial self-attention map, we produce the second component of the ReplaceBlock, i.e., the replacement region sampling map (\textbf{RR-SM}). The RR-SM is used to obtain the contiguous regions in the original image feature map that need to be replaced (The discriminative regions in the deep feature map of the original image have a higher probability of being replaced).

We summarize our contribution as follows:
\begin{itemize}
\item We propose a novel regularization method, which can use the background information to effectively simulate the deep feature map of the occluded image, improving the generalization ability of the model.
\item Our method replaces parts of the most discriminative region, which can help the model capture the full extent of the object, resulting in cleaner background-only images.
\item Compared with DropBlock, ReplaceBlock significantly improves the performance in image classification and fine-grained image recognition tasks.
\end{itemize}

\section{Related Work}

This section summarizes two deep learning topics relevant to this work: (1) dropout-based regularization method and (2) attention mechanisms.
\subsection{Dropout-based regularization method}

In an attempt to apply a structured form of Dropout to the convolutional layer, Tompson and Jonathan \cite{tompson2015efficient} proposed SpatialDropout that randomly drops partial channels of a feature map, rather than dropping each pixel. Cutout \cite{devries2017improved} drops contiguous regions of input images instead of individual pixels in the input layer of CNNs. This method induces the network to better utilize the contextual information of the image, rather than relying on a small set of speciﬁc features. Meanwhile, CamDrop \cite{10.1145/3357384.3357999} selectively abandons some specific spatial regions in predominating visual patterns by considering the intensity of class activation mapping (CAM) all around to improve the efficiency of abandoning. DropBlock \cite{ghiasi2018dropblock} generalizes Cutout by applying Cutout at every feature map in convolutional networks. CutMix \cite{Yun_2019_ICCV} proposes an augmentation strategy: patches are cut and pasted among training images where the ground truth labels are also mixed proportionally to the area of the patches. Different from CutMix, which mixes two images and their labels to augment the input data, ReplaceBlock uses background information to simulate feature maps of occluded images.

\subsection{Attention mechanism}
Attention mechanism \cite{tsotsos1990analyzing} have been proven helpful in a variety of computer vision tasks, such as image classiﬁcation \cite{hu2018gather} and face recognition \cite{hu2018squeeze}. The squeeze-and-excitation module \cite{hu2018squeeze} was introduced to exploit the channel-interdependencies. The module has been successful in face recognition tasks. Bottleneck Attention Module (BAM) \cite{park2018bam} and Convolutional Block Attention Module (CBAM) \cite{woo2018cbam} further extends this idea by introducing spatial information encoding via additional(auxiliary) large-size convolution kernels.

Attention mechanisms are also widely used in WSOL. WSOL is an alternative cheaper way to identify the object location in a given image by solely using the image-level supervision \cite{zhou2016learning,choe2019attention,yin2021dual}. CAM \cite{zhou2016learning} revisit the global average pooling layer, and shed light on how it explicitly enables the CNN to have remarkable localization ability. ADL \cite{choe2019attention} propose an Attention-based Dropout Layer, which utilizes the self-attention mechanism to hide the most discriminative part from the model for capturing the integral extent of object. However, the key difference from them is that we locate the target object with the aim of obtaining an image with only background information to establish the basis for the replacement operation.

\section{Proposed Approach}

\begin{figure*}[t]
\centering
\includegraphics[width=0.8\linewidth]{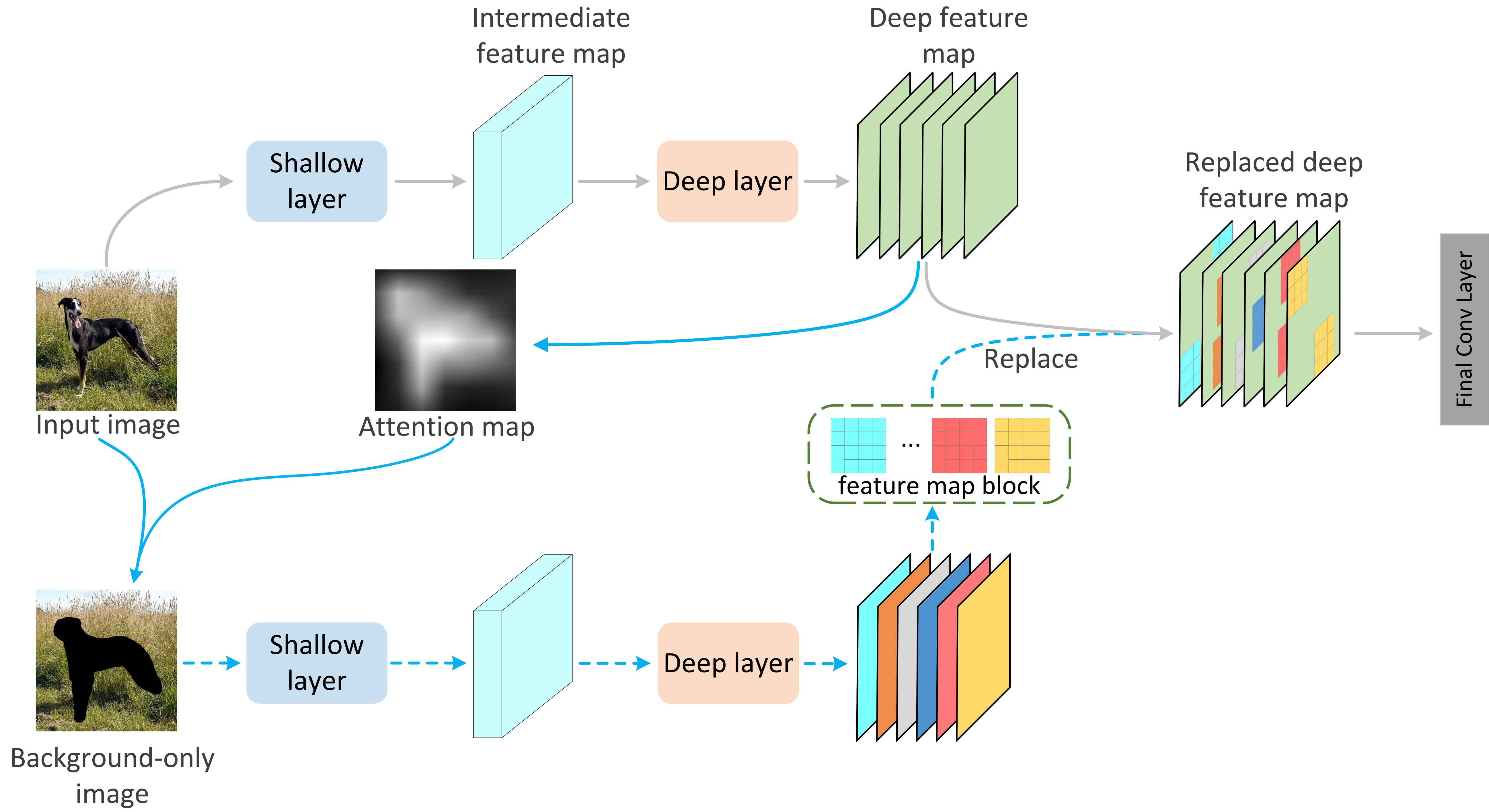} 
\caption{ReplaceBlock diagram. The blue arrows indicate the method of generating the background-only image and the background information feature map. The ``Attention map'' is generated from the deep feature map by CAM. Note that ``Shallow layer'' and ``Deep layer'' in the figure share parameters in both calculations.}
\label{MethodBlockDiagram}
\end{figure*}

\subsection{ReplaceBlock}

Here, we take replacing the input feature map of the final convolutional layer as an example to introduce ReplaceBlock, as shown in Figure \ref{MethodBlockDiagram}. Let ${{G}_{bb}}(\centerdot )$ be all the convolutional layers of the model excepting the final conv layer, and ${{G}_{final}}(\centerdot )$ be the final conv layer of the model. Given the input image $x$, we have its feature maps:
\begin{equation}
{{\mathrm{F}}_{pf}}={{G}_{bb}}(x)
  \label{eq:1}
\end{equation}
and
\begin{equation}
 {{\mathrm{F}}_{ff}}={{G}_{final}}({{\mathrm{F}}_{pf}})
  \label{eq:2}
\end{equation}
where $\mathrm{F}\in {{\mathrm{R}}^{H\times W\times C}}$ be a feature map. Note that $C$ denotes the channel number, $H$ and $W$ are the height and the width of the feature map, respectively. For simplicity, the mini-batch dimension is omitted in this notation.

\begin{figure}[h]
\centering
\includegraphics[width=0.8\linewidth]{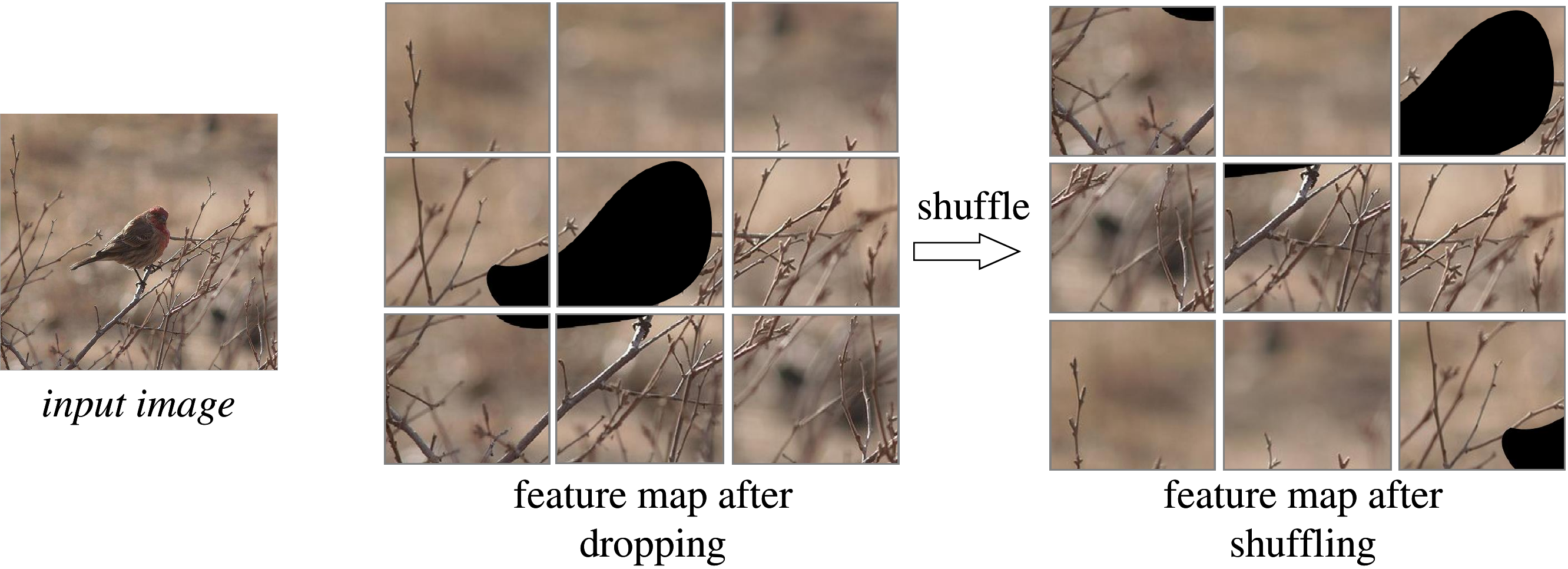}
\caption{Random shuffling increased the randomness of the background and making it possible for the ReplaceBlock to extract the background of any position. Random shuffling is performed at the deep feature map level.}.
\label{shuffe}
\end{figure}

To obtain the TC-DM, we perform CAM \cite{zhou2016learning} on the ${{\mathrm{F}}_{pf}}$ to derive the 2D spatial self-attention map ${{\mathrm{M}}_{mask}}\in {{\mathrm{R}}^{H\times W}}$ . We observed that the ${{\mathrm{M}}_{mask}}$ usually highlights the most discriminative regions of the target object and suppresses less discriminative ones. We use a simple thresholding technique to segment the self-attention map to obtain the TC-DM: ${{\mathrm{B}}_{img}}$. We first set a drop threshold value that is 20\% of the max value of the ${{\mathrm{M}}_{mask}}$. Then, we generate the TC-DM: ${\mathrm{B}}_{img}\in {{\{0,1\}}^{H\times W}}$ by setting each pixel to 0 if it is larger than the drop threshold, and 1 if it is smaller. That is, the drop mask has 0 for the region to be recognized and 1 for the background region. The drop mask is applied to the input image by element-wise multiplication. In this way, we can hide the region of the target object in the input image. Calculation of the feature map for the irrelevant objects and background:
\begin{equation}
{{\mathrm{F}}_{BG\_pf}}={{G}_{bb}}({{\mathrm{B}}_{img}}\odot x)
  \label{eq:3}
\end{equation}
where $\odot $ is the general element-wise multiplication, ${{\mathrm{F}}_{BG\_pf}}$ denotes the feature map of the image with only irrelevant objects and background. To use the background information more effectively, we randomly shuffle ${{\mathrm{F}}_{BG\_pf}}$ in the spatial dimension, as shown in Figure \ref{shuffe}. Using the random shuffle strategy allows the ReplaceBlock to fetch the background at any position. Note that ${{\mathrm{F}}_{BG\_pf}}$ and ${{\mathrm{F}}_{pf}}$ are calculated from the same ${{G}_{bb}}(\centerdot )$ in each forward propagation. 

Then, we use the ${{\mathrm{M}}_{mask}}$ sampling to get ${{\mathrm{M}}_{sample}}:{{\mathrm{M}}_{sample}}\sim B({{\mathrm{M}}_{mask}})$. Using ${{\mathrm{M}}_{mask}}$ makes it more likely to drop regions with stronger discriminability. Because there is a strong correlation between the neighboring pixels on the convolutional layers, and these adjacent pixels have similar information, we use a similar approach to DropBlock to obtain a structured form of drop mask ${{\mathrm{B}}_{feature}}$ from ${{\mathrm{M}}_{sample}}$. 
\begin{figure}[h]
\centering
\includegraphics[width=0.8\linewidth]{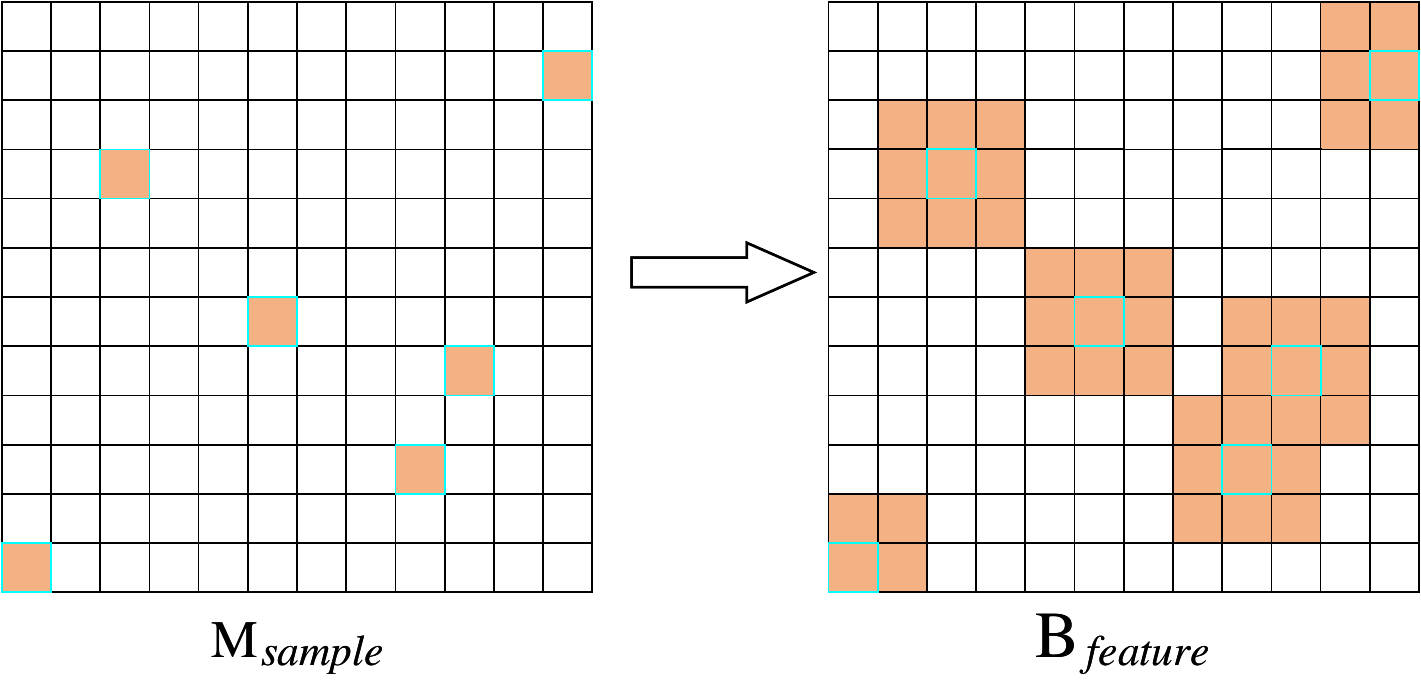}
\caption{To adapt the convolutional layer, we obtain a structured form of drop mask.}.
\label{structuredDrop}
\end{figure}
${{\mathrm{B}}_{feature}}$ drops the contiguous regions of feature maps rather than the individual pixels, as illustrated in Figure \ref{structuredDrop}. The contiguous regions of the original feature map ${{\mathrm{F}}_{pf}}$ indicated by ${{\mathrm{B}}_{feature}}$ are replaced by the corresponding regions of the background feature map ${{\mathrm{F}}_{BG\_pf}}$, formally as :

\begin{equation} 
\begin{aligned}
& {{\mathrm{F}}_{replace\_pf}}={{\mathrm{F}}_{pf}}\odot {{\mathrm{B}}_{feature}}+{{\mathrm{F}}_{BG\_pf}}\odot (\text{\textbf{1}}-{{\mathrm{B}}_{feature}}) \\
\end{aligned}
 \label{eq:4} 
\end{equation}
where $\text{\textbf{1}}$ is a binary mask ﬁlled with ones. Finally, the output feature of the final conv layer is as follows:
\begin{equation}
{{\mathrm{F}}_{replace\_ff}}={{G}_{final}}({{\mathrm{F}}_{replace\_pf}})
  \label{eq:3}
\end{equation}

In this way, we replace a small portion of the most discriminative regions with feature maps of irrelevant objects and background, encouraging the model to adapt to the situation where the most discriminative regions of the image are partially occluded.

\subsection{Background-only image after being masked}

ReplaceBlock inputs the masked image (${{\mathrm{B}}_{img}}\odot x$) directly into the model to extract features. To make more effective use of background information, we have tried to use image inpainting technology to repair ${{\mathrm{B}}_{img}}\odot x$ and obtain a background-only image without black regions to get the complete background information. The experiment found that the training time cost is too large to obtain a credible repair result, and it is not feasible. How to better process ${{\mathrm{B}}_{img}}\odot x$ is the current limitation of ReplaceBlock, and it is also one of the concerns of our next step research.

\section{Experiments}

In this section, we first describe our experiment dataset and experiment settings, then conduct an ablation experiment to demonstrate the relation of the background information to the proposed ReplaceBlock to the performance. Next, we compare our approach with some of the most popular current regularization methods. Finally, we visualize the self-attention map obtained by our proposed method and the vanilla method.

\subsection{Experiment Settings}
\noindent \textbf{Dataset}. We evaluate the performance of our method using the top-1 accuracy of Caltech -256 \cite{griffin2007caltech} and Mini-ImageNet \cite{krizhevsky2012imagenet}. Caltech256 has 256 classes with more than 80 images in each class. Mini-ImageNet contains a total of 60,000 images from 100 classes. To validate the proposed ReplaceBlock on the fine-grained image classification (FGVC) task, we test it on CUB2011 \cite{wah2011caltech} and Cars-196 \cite{krause2015fine}. CUB-2011 is a bird dataset that consists of 200 bird species and 11788 images. Cars-196 consists of 196 car classes and 16185 images.

\noindent \textbf{Implementation details}. We use the PyTorch toolbox \cite{paszke2019pytorch} to implement all our experiments. During training, we used the standard SGD optimizer with a momentum of 0.9 to train all the models. The weight decay is set to 4E-5 always. The cosine learning schedule \cite{he2019bag} with an initial learning rate of 0.01 is adopted. All experiments are conducted on an NVIDIA GeForce GTX 1080Ti (11 GB) GPU. We use ResNet-50 as the backbone. The images are cropped to 224 × 224 as the input. Random horizontal flipping and random cropping are used for data augmentation. 

\noindent \textbf{Where to apply ReplaceBlock.} For a fair comparison with DropBlock, we follow the setting of DropBlock \cite{ghiasi2018dropblock}, applying ReplaceBlock in the last two Groups of ResNet-50. In a Group, apply ReplaceBlock after both convolution layers and skip connections.

Generating ${{\mathrm{B}}_{feature}}$ requires two hyperparameters, \emph{keep-prob} and \emph{block-size}. \emph{keep-prob} controls the drop rate, and \emph{block-size} controls the size of the dropped region. In this paper, we refer to the setting of DropBlock with \emph{keep-prob}=0.9 and \emph{block-size}=3.

\subsection{Ablation Studies}

\begin{figure}[t]
 \centering
\includegraphics[width=0.8\linewidth]{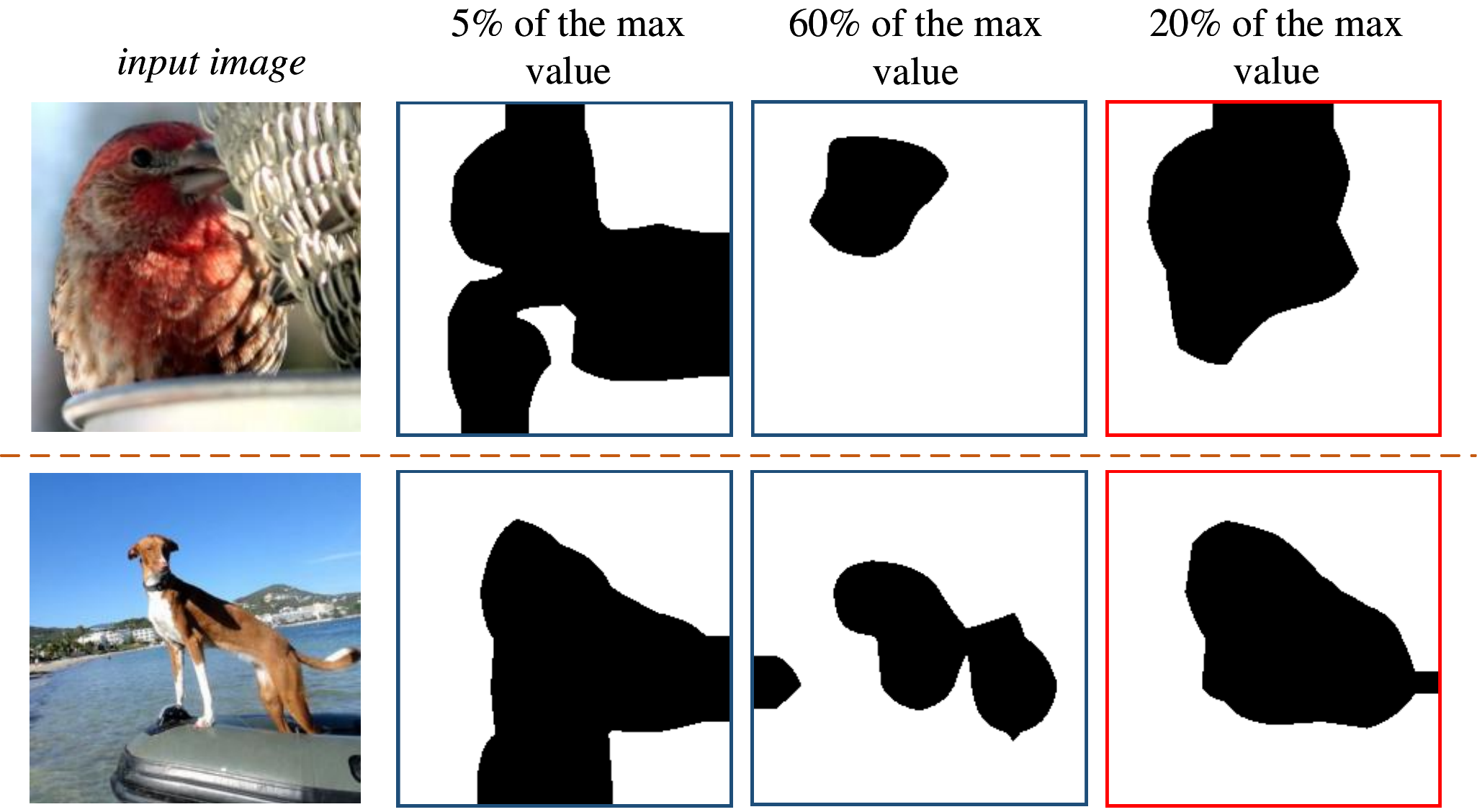}
\caption{By setting different thresholds for the CAM, different TC-DMs can be obtained.}.
\label{differentMaxValue}
\end{figure}

To demonstrate the effect of different background information on the performance of the proposed ReplaceBlock, we set different thresholds to the CAM to obtain TC-DM, resulting in different background-only images. First, we visualize the image masks obtained by different thresholds in Figure \ref{differentMaxValue}. The region dropped by CAM(5\%) includes not only the object to be recognized, but also other objects in the image, as shown in the second column of Figure \ref{differentMaxValue}; although the object to be recognized is the dog, the region of the boat in the image is also dropped. CAM(60\%) only drops the most discriminative regions in image, which has the problem of under-dropping. From Figure \ref{differentMaxValue}, it can be seen that the best image mask is obtained by CAM(20\%).

\begin{table}[tb]
   \centering
   \caption{Top-1 accuracy (\%) for ReplaceBlock with different background information, on Mini-ImageNet and Caltech-256. All models were trained in the same experimental settings.}
   \begin{tabular}{l|c|c}
      \hline 
      \small{Method}                   & \small{Mini-ImageNet}     & \small{Caltech-256} \\
      \hline 
      \hline 
      \small{ResNet50 (Baseline) }     & 80.13          & 66.21 \\
      \hline 
      \small{ReplaceBlock (CAM(60\%))}            & 82.52          & 68.61 \\
      \small{ReplaceBlock (CAM(5\%))}  & 82.50 & 68.65 \\
     \small{\textbf{ReplaceBlock (CAM(20\%))}}          & \textbf{83.03}          & \textbf{69.75} \\
      \hline 
   \end{tabular} \\
   \label{ImageNetAndCaltech}
\end{table}

Then, we use ResNet-50 as a baseline to test the performance variation of the proposed ReplaceBlock method when different image masks are applied. In this experiment, we adopt three image mask obtaining approaches described above. As shown in Table \ref{ImageNetAndCaltech}, the model using CAM(20\%) has the best performance, which corresponds to the masked image shown in Figure \ref{differentMaxValue}. This indicates that the performance of ReplaceBlock was related to the background information used as a replacement and that a cleaner background image lead to a better performance improvement. The ReplaceBlock is the same as the DropBlock when using a background image with all pixel values of 0.

To demonstrate the impact of RR-SM on the performance of ReplaceBlock, we obtain drop masks by random sampling and RR-SM sampling, respectively. As shown in Table \ref{sampling method}, using RR-SM gets better results, but the improvement is smaller. This may be because when there are enough training epochs, the randomly generated drop mask also contains a large number of samples that drop the more discriminative regions.

\begin{table}[h]
   \centering
   \caption{Results of different sampling methods on Mini-ImageNet and Caltech-256 (Top-1 acc, \%).}
   \begin{tabular}{l|c|c}
      \hline 
      Method                   & \footnotesize{Mini-ImageNet}     & \footnotesize{Caltech-256} \\
      \hline 
      \hline 
      \scriptsize{ResNet50 (Baseline)}      & 80.13          & 66.21 \\
      \hline 
      \scriptsize{ReplaceBlock + random sampling}             & 82.89          & 69.57 \\
      \scriptsize{\textbf{ReplaceBlock + RR-SM}}          & \textbf{83.03}          & \textbf{69.75} \\
      \hline 
   \end{tabular} \\
   \label{sampling method}
\end{table}

To explore whether ReplaceBlock leads to 'background bias' (the model does not see the target object, but use backgrounds in classification), we set up an experiment to train the model in two ways: (1) applying the ReplaceBlock at each training step, and (2) applying the ReplaceBlock alternately (the current training step applies the ReplaceBlock, and the next training step does not apply the ReplaceBlock). As shown in Table \ref{alternate application}, the performance of applying ReplaceBlock all the time is much better than that of applying ReplaceBlock alternately. We think there are two possible reasons: (1) the drop rate set by ReplaceBlock is small (10\%), the discriminative region of the feature map is still dominant, (2) Different images and different positions of the same image both contain different backgrounds, making it difficult to overfit the model to a particular background, so ReplaceBlock can simulate the feature map of the occluded image well without causing 'background bias'.

\begin{table}[h]
   \centering
   \caption{Results of alternate application of replaceblock and all-time application of replaceblock on Mini-ImageNet and Caltech-256. (Top-1 acc, \%).}
   \begin{tabular}{l|c|c}
      \hline 
      Method                   & Mini-ImageNet     & Caltech-256 \\
      \hline 
      \hline 
      ResNet50 (Baseline)      & 80.13          & 66.21 \\
      \hline 
      \small{ReplaceBlock (alternate)}             & 81.73          & 68.25 \\
      \small{\textbf{ReplaceBlock (all-time)}}          & \textbf{83.03}          & \textbf{69.75} \\
      \hline 
   \end{tabular} \\
   \label{alternate application}
\end{table}

\subsection{Comparison with Other Methods}

\subsubsection{Mini-ImageNet and Caltech-256.}

\begin{table}[h]
   \centering
   \caption{Top-1 and Top-5 accuracies (\%) for ResNet-50 model with different regularization methods, on Mini-ImageNet datasets. The ReplaceBlock module follows the best settings in the ablation studies, the other regularization methods follow the settings in their papers.}
   \begin{tabular}{l|c|c}
      \hline 
      Method                   & Top-1 acc.     & Top-5 acc. \\
      \hline 
      \hline 
      ResNet50 (Baseline)      & 80.13          & 93.53 \\
      \hline 
      Dropout (kp=0.7)             & 80.41          & 93.91 \\
      DropPath (kp=0.9)&	81.21  &	93.88\\
      SpatialDropout (kp=0.9) & 81.53  & 93.97 \\
      Cutout &	80.63 &	93.62\\
      Cutmix &	82.47 &	93.95\\
      DropBlock (kp=0.9) &82.43  &94.56\\
     \textbf{ReplaceBlock (kp=0.9)}& \textbf{83.03}  & \textbf{94.83}\\
      \hline 
   \end{tabular} \\
   \label{Mini-ImageNet (Top-1 and Top-5)}
\end{table}

We compare our ReplaceBlock with other regularization methods used in CNNs, including the widely used Cutout \cite{devries2017improved}, DropPath \cite{zheng2020pac}, CutMix \cite{Yun_2019_ICCV}, SpatialDropout \cite{tompson2015efficient} and DropBlock \cite{ghiasi2018dropblock}. As can be seen from Table \ref{Mini-ImageNet (Top-1 and Top-5)} and Table  \ref{Caltech-256 (Top-1 and Top-5)}, adding the DropBlock has already improved the classiﬁcation accuracy by more than 2.3\%. The Cutout has a lower performance than DropBlock, which indicates that dropping at the feature map is better than dropping directly on the input image. However, when the proposed ReplaceBlock is considered, we achieve the best results. 

\begin{table}[h]
   \centering
   \caption{Results on Caltech-256 (Top-1 and Top-5 acc, \%).}
   \begin{tabular}{l|c|c}
      \hline 
      Method                   & Top-1 acc.     & Top-5 acc. \\
      \hline 
      \hline 
      ResNet50 (Baseline)      & 66.21          & 81.39 \\
      \hline 
      Dropout (kp=0.7)             & 67.55          & 82.34 \\
      DropPath (kp=0.9)&	68.36  &	83.76\\
     SpatialDropout (kp=0.9) & 68.50  & 84.07 \\
       Cutout &	68.15 &	83.67\\
       Cutmix &	69.27 &	84.25\\
       DropBlock (kp=0.9) &69.39  &84.19\\
      \textbf{ReplaceBlock (kp=0.9)}& \textbf{69.75}  & \textbf{84.82}\\
      \hline 
   \end{tabular} \\
   \label{Caltech-256 (Top-1 and Top-5)}
\end{table}

We argue that the advantages of the proposed ReplaceBlock over DropBlock are two-fold. First, the DropBlock directly drops some regions of the feature map, which cannot well simulate the occlusion in the image. Because the features in the occluded region of the image are not dropped directly, but the distraction features of a certain object appears. However, our ReplaceBlock replaces some regions in the feature map of the original image using the features of background, which can effectively simulate the feature map of the occluded image. Second, DropBlock drops the regions in feature map randomly, while our ReplaceBlock uses self-attention map to produce replacement regions. This enables our ReplaceBlock to alleviate the over-fitting of the model more efficiently.

\subsubsection{CUB-2011 and Cars-196.}

To validate the proposed ReplaceBlock on the FGVC task, we test it on two datasets, CUB-2011 and Cars-196. As the results are shown in Table \ref{CUB-2011 and Cars-196}, we can draw a similar conclusion as Table \ref{Mini-ImageNet (Top-1 and Top-5)} and Table \ref{Caltech-256 (Top-1 and Top-5)}. ReplaceBlock achieves the best results on these two FGVC datasets. Compared with DropBlock, our method still gains 0.84\% and 0.49\%, respectively.

\begin{table}[h]
   \centering
   \caption{Top-1 accuracy (\%) for ResNet-50 model with different regularization methods, on FGVC tasks CUB-2011 and Cars-196. The ReplaceBlock module follows the best settings in the ablation studies, the other regularization methods follow the settings in their papers.}
   \begin{tabular}{l|c|c}
      \hline 
      Method                   & CUB-2011     & Cars-196 \\
      \hline 
      \hline 
      ResNet50 (Baseline)      & 80.96          & 89.56 \\
      \hline 
       Dropout (kp=0.7)             & 80.97          & 89.77 \\
       DropPath (kp=0.9)&	81.11  &	90.27\\
      SpatialDropout (kp=0.9) & 81.26  & 90.06 \\
      Cutout &	80.98 &	89.75\\
       Cutmix &	81.27 &	91.23\\
       DropBlock (kp=0.9) &81.41  &91.68\\
      \textbf{ReplaceBlock (kp=0.9)}& \textbf{82.25}  & \textbf{92.17}\\
      \hline 
   \end{tabular} \\
   \label{CUB-2011 and Cars-196}
\end{table}

\begin{figure*}[t]
\centering
\includegraphics[width=1.0\linewidth]{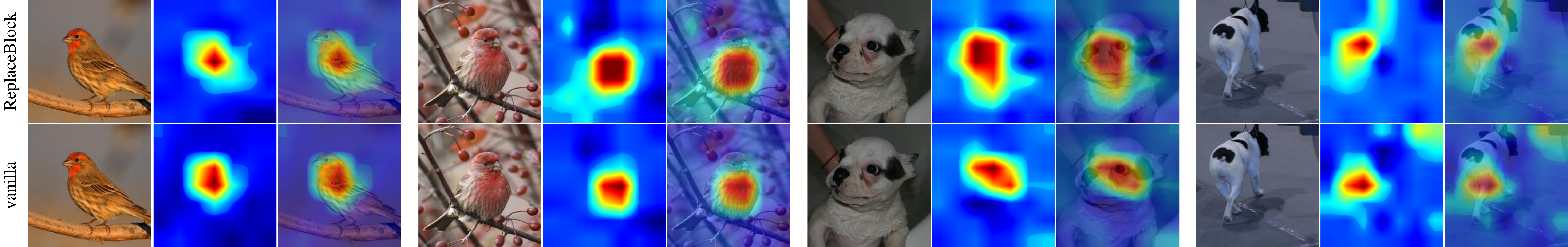} 
\caption{Qualitative evaluation results of ReplaceBlock on Mini-ImageNet. We compared our method and the vanilla model. The left image in each ﬁgure is input image. The middle image is heatmap and the right image shows the overlap between the input image and the heatmap.}
\label{heatmap}
\end{figure*}

\subsection{Visualize the heatmap}

To validate whether ReplaceBlock can help the model capture a complete target object, we show the heatmaps obtained by different methods. Figure \ref{heatmap} visualizes the heatmaps of our method and the vanilla model on the Mini-ImageNet datasets. From the results, we consistently observe that model with ReplaceBlock captures more complete target object than the vanilla model. The vanilla model usually captures only the most discriminative parts. For example, as seen from the left-most sample in the Figure \ref{heatmap}, the heatmap extracted from the vanilla model only highlight the neck of birds. Contrarily, the model with ReplaceBlock covers not only the neck, but also the entire part of the bird, from neck to tail. In addition, from the third sample in Figure \ref{heatmap}, the vanilla model focuses only on the eyes of dogs, whereas the model with ReplaceBlock localizes the entire region of the dog.

Based on the above analysis, ReplaceBlock can help the model capture the full extent of target object and obtain a cleaner background-only image to improve the model's efficiency in utilizing background information.

\section{Conclusions}

In this paper, we present a novel regularization method for CNNs, named as ReplaceBlock. ReplaceBlock is a form of structured replacement that utilizes irrelevant background information to replace spatially correlated region. ReplaceBlock is an auxiliary module which is applied only during training stage. And in the testing stage, ReplaceBlock is deactivated. Our ReplaceBlock inherits the advantages of DropBlock in structuring the erasure information, but ReplaceBlock uses a replacement operation instead of direct dropout to better simulate the case of the occluded image, resulting in a more diverse feature map. Meanwhile, ReplaceBlock can induce the CNN classifier to learn the entire extent of the object and obtain cleaner background-only images, which provides a guarantee for using background information. Experiments in Mini-ImageNet classiﬁcation and fine-grained image classification demonstrate the effectiveness of our ReplaceBlock. We hope that our work will promote the study of regularization methods for the application of irrelevant background information.

\bibliographystyle{named}
\bibliography{ijcai22}

\end{document}